\documentclass[10pt, a4paper]{article}

\usepackage{lrec-coling2024} 
\usepackage{orcidlink}
\RequirePackage{pgfplots}
\RequirePackage{booktabs}
\RequirePackage{mathtools}
\RequirePackage[english]{babel}
\RequirePackage[autostyle, english = american]{csquotes}
\RequirePackage{mwe}
\RequirePackage{graphicx}
\RequirePackage{caption}
\RequirePackage{subcaption}

\title{Improving noisy student training for low-resource languages in End-to-End ASR using CycleGAN and inter-domain losses}

\name{Chia-Yu Li and Ngoc Thang Vu} 

\address{Institute for Natural Language Processing (IMS), University of Stuttgart \\
         Pfaffenwaldring 5b, 70569 Stuttgart, Germany\\
         licu@ims.uni-stuttgart.de, thang.vu@ims.uni-stuttgart.de\\}

\abstract{
Training a semi-supervised end-to-end speech recognition system using noisy student training has significantly improved performance. However, this approach requires a substantial amount of paired speech-text and unlabeled speech, which is costly for low-resource languages. Therefore, this paper considers a more extreme case of semi-supervised end-to-end automatic speech recognition where there are limited paired speech-text, unlabeled speech (less than five hours), and abundant external text. Firstly, we observe improved performance by training the model using our previous work on semi-supervised learning \enquote{CycleGAN and inter-domain losses} solely with external text. Secondly, we enhance \enquote{CycleGAN and inter-domain losses} by incorporating automatic hyperparameter tuning, calling \enquote{enhanced CycleGAN inter-domain losses.} Thirdly, we integrate it into the noisy student training approach pipeline for low-resource scenarios. Our experimental results, conducted on six non-English languages from Voxforge and Common Voice, show a 20\% word error rate reduction compared to the baseline teacher model and a 10\% word error rate reduction compared to the baseline best student model, highlighting the significant improvements achieved through our proposed method. 
 \\ \newline \Keywords{speech recognition, low resource, semi-supervised training, CycleGAN, noisy student training} }

\begin{document}

\maketitleabstract

\section{Introduction}
\begin{figure*}[htbp]
  \centering
  \begin{tabular}[c]{ccc}
    \begin{subfigure}[b]{.3\textwidth}
      \includegraphics[width=\textwidth]{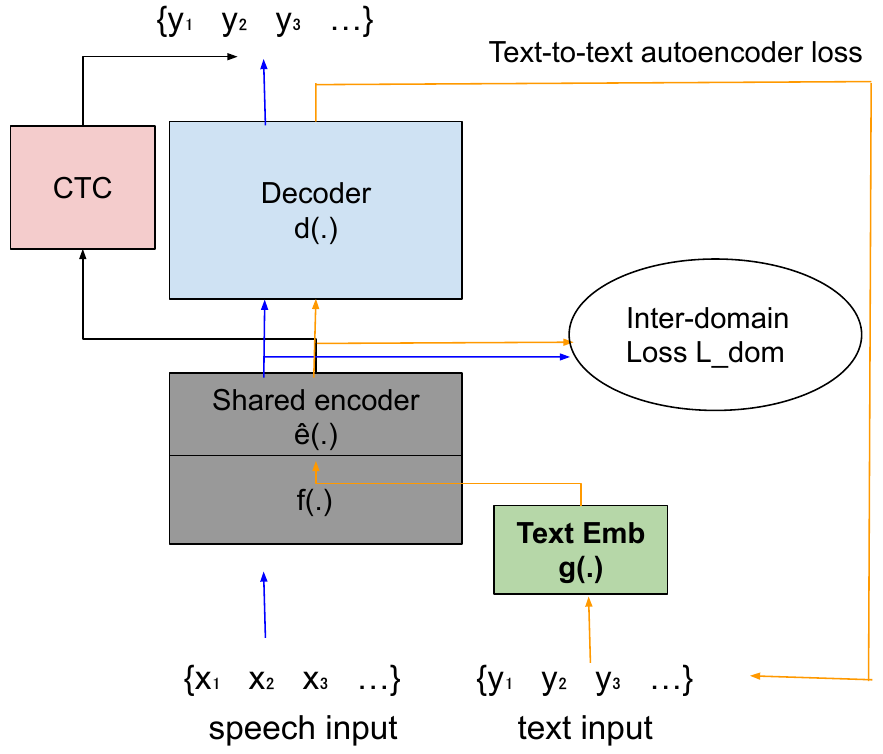}
      \caption{The architecture of semi-supervised E2E ASR.}
      \label{fig:architecture-semi-supervised-e2e-asr}
    \end{subfigure}&
    \begin{subfigure}[b]{0.3\textwidth}
      \includegraphics[width=\textwidth]{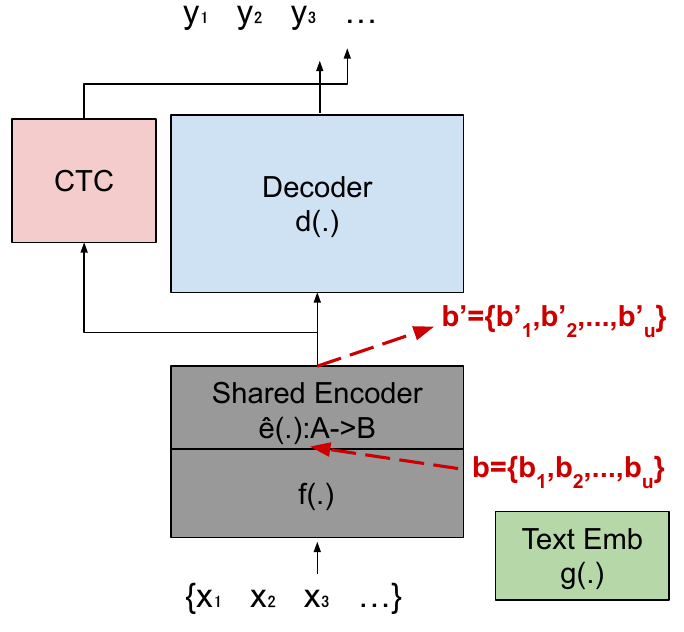}
      \caption{The identity mapping loss. Note that $b$ is the representations from encode speech or text.}
      \label{fig:illustration-idt-loss}
    \end{subfigure}&
    \begin{subfigure}[b]{0.3\textwidth}
      \includegraphics[width=\textwidth]{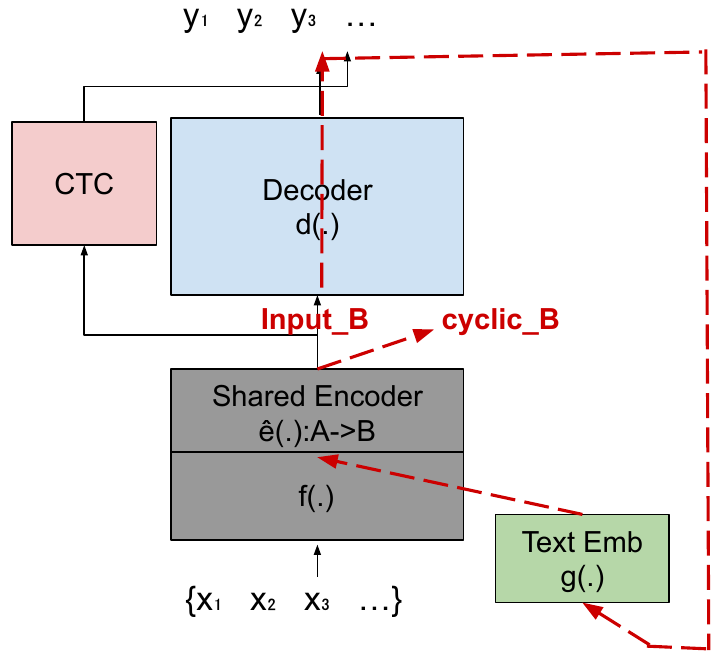}
      \caption{The cycle-consistent inter-domain loss.}
      \label{fig:illustration-cyc-dom-loss}
    \end{subfigure}
  \end{tabular}
  \caption{The framework of CycleGAN and inter-domain losses \cite{CY_cycleGAN-inter-domain-losses}.}\label{fig:animals}
\end{figure*}
Over the last decade, there has been a significant improvement in the performance of speech and language processing technologies, with an increasing number of systems being deployed across multiple languages and applications. However, the majority of these efforts have been focused on a limited set of languages. Given that there are over 6,900 languages worldwide, the biggest challenge today is to quickly and cost-effectively transfer speech processing systems to new languages with minimal manual effort. In the field of automatic speech recognition (ASR), semi-supervised end-to-end (E2E) can be applied to reduce the amount of annotated data. Two prominent approaches include consistency-based and iterative self-training-based methods. The consistency-based method focuses on enhancing the model by improving the representation of input through training a separate task \cite{tjandra_listen_while_speaking,hayashi_back_translation_DA,Renduchintala_multi-modal_DA,karital_inter-domain-loss,hsu_domain_invariant_features,chung_speech2vec,Chorowski_wavnet_autoencoders,hori_cycle-consistency-training,schneider_wav2vec,Baevski_vq-wav2vec,Ling_deep_contextualized_acoustic_rep}. The iterative self-training technique utilizes augmentation to improve the overall network performance \cite{Zavaliagkos_utilizing_untranscribed_training_data,Novotney_self-training-low-resource,Thomas_DNN_feature_and_semi-supervised_training,parthasarathi_lessons_from_building_AM_with_a_million_hours_of_speech,Li_weak-distillation_semi-supervised_training,Kahn_self-training_E2E,Synnaeve_E2E_ASR_semi-supervised_modern_architecture,hsu_self-supervised_ASR_via_local_prior_matching}. Among the various techniques, a widely recognized approach known as noisy student training (NST) has emerged. NST is an iterative self-training method that leverages unlabeled data to enhance accuracy, particularly in the domains of image classification and machine translation \cite{Xie_noisy_self_training_image_classification}. Park et al. adapted and improved NST by employing techniques such as SpecAugment \cite{specAug1,specAug2} and incorporating shallow fusion with a language model (LM) into the teacher network. Additionally, they introduced a normalized filtering score that aids in generating enhanced transcripts for training the student network \cite{Park_NST}. The results demonstrate significant performance on Librispeech \cite{librispeech} and LibriLight \cite{librilight}.

Although NST is simple and effective, it depends on a substantial quantity of paired speech-text to train a teacher model, which is used for labeling the unlabeled speech data that the student model could train on. For low-resource languages, the paired speech-text is expensive. There are techniques that can be explored to address this limitation. One approach is to leverage pre-trained models, such as wav2vec \cite{schneider_wav2vec}, where leverages transfer learning to learn contextual representations from a large corpus of unlabeled speech data. The model can then be fine-tuned for the target domain using unlabeled speech data from the same target domain. However, this approach still requires a reasonable quantity of speech data, which is still expensive in low-resource scenario. Besides, this technique requires multi-stage tuning processing which introduces computational cost. How to improve inexpensively the teacher model in NST remains a key challenge especially in language with very small data.

Our previous work \enquote{cycle-consistent generative adversarial networks (CycleGAN) and inter-domain losses}, which is the dissimilarity between the intermediate representations of encoded speech and its hypothesis \cite{CY_cycleGAN-inter-domain-losses}, was proposed for semi-supervised E2E ASR. The architecture is shown in \autoref{fig:architecture-semi-supervised-e2e-asr}. CycleGAN and inter-domain losses (CID) encourage the model to learn the common representations from the speech and text. With the advantage of this structure allowing speech and text input, we observe that training a model by CID with small paired speech-text and additional external text (without additional speech) can still improve the ASR performance. Therefore, we propose leveraging it into the training pipeline of NST to enhance the teacher model solely using a large amount of external text. Subsequently, the improved teacher model generates better labels for the unlabeled speech, which the student model can train on.

In this paper, we make several contributions in the following aspects: Firstly, we observe that training a model by CID \cite{CY_cycleGAN-inter-domain-losses} with lots of external text significantly boosts performance (\autoref{sec:better_cyclegan-inter-domain-losses-with-text-data}); Secondly, we enhance CID by incorporating automatic
hyperparameter tuning, calling enhanced CID (\autoref{sec:better_cyclegan-inter-domain-losses}); Thirdly, we improve the NST training pipeline for low-resource scenarios by boosting the teacher model using enhanced CID (\autoref{sec:cyclegan-inter-domain-losses_and_NST}); Fourthly, we evaluate our method on six languages on the Voxforge and Common Voice (\autoref{sec:experimental_setup} and \autoref{sec:result}). The results demonstrate that our proposed approach achieves a 20\% word error rate reduction (WERR) compared to the baseline (NST) teacher model, and a 10\% WERR compared to the baseline student model for most languages. Notably, the improvement of teacher model is accomplished without the need for additional speech data. Lastly, we provide an analysis of the recognition output and cherry-pick hypothesis (\autoref{sec:analysis}).

For the sake of simplicity, throughout the rest of this paper, we use the term \enquote{paired data} to refer to \enquote{paired speech-text,} the term \enquote{unpaired data} to refer to \enquote{unpaired speech-text,} the term \enquote{CID} to refer to the \enquote{CycleGAN and inter-domain} approach, and our proposed NST pipeline designed for low-resource using CID is denoted as \enquote{cNST}.

\section{Method}
\begin{table*}[htbp]
\small
    \centering
    \begin{tabular}{lcccc}
    \toprule
    \textbf{Model} &\textbf{paired data}& \textbf{unpaired text}&  \textbf{without LM} & \textbf{ with LM}\\
          && (\#lines) &WER(\%)&WER(\%)\\\midrule\midrule
    Initial model ($M_0$) & Voxforge German (5 hrs.)&0                     & 63.6 &63.1 \\
    CID model ($M_1$)& Voxforge German (5 hrs.)& 10K \cite{leipzig_corpora}& 38.6 &36.3\\
                    & Voxforge German (5 hrs.)& 100K \cite{leipzig_corpora}& 31.2 &29.4\\
                    & Voxforge German (5 hrs.)& 300K \cite{leipzig_corpora}& \textbf{30.8} & \textbf{29.1}\\\hline
    \end{tabular}
    \caption{WERs on the Voxforg German test set. Note that  the initial model is trained by supervised objective in \autoref{eq:loss_pair} with five-hour Voxforg German train data, and the CID model ($M_1$) is trained with same five-hour Voxforg German train data and external text from Leipzig corpus \cite{leipzig_corpora} via semi-supervised objective in \autoref{eq:semie2e}.}
    \label{tab:result1}
\end{table*}
\label{sec:method}
\subsection{CycleGAN and inter-domain losses (CID)}
Figure \ref{fig:architecture-semi-supervised-e2e-asr} shows the CID architecture, which is based on semi-supervised E2E speech recognition and joint CTC-attention E2E \cite{jointlyc2catt,hybride2e,karital_inter-domain-loss}. The encoder is $e=\hat e \circ f$ when the input is speech. If the input is text, the encoder is the composition of text embedding $g(.)$ and the share encoder $\hat e$. i.e., $\hat e \circ g$. The model is trained by jointly CTC-attention objective on paired data $S=\{X, Y\}$ and by CID on unpaired data $U=\{X', Y'\}$ simultaneously. The objective is as follows \cite{karital_inter-domain-loss,CY_cycleGAN-inter-domain-losses},
\begin{equation}
    \mathcal{L}=\alpha \mathcal{L}_{pair}(e,d,S)+ (1-\alpha)\mathcal{L}_{unpair}(f,g,\hat e,d,U)\label{eq:semie2e}
\end{equation}
where the supervised ratio $\alpha$ is a tunable parameter.

The supervised objective is negative log likelihood of the ground-truth $y$ given the encoded speech $e(x)$ \cite{hybride2e}:
\begin{align}
\label{eq:loss_pair}
\begin{split}
    \mathcal{L}_{pair}(e,d,S)&=-\sum_{(x,y)\in S} \log d(e(x))\\
    &=-\sum_{(x,y)\in S} \log \prod_{t=1}^{|y|} \Pr(y_{t}|y_{t-1},e(x))
\end{split}
\end{align}

The unsupervised objective CID consists of the identity mapping loss, the cycle-consistent inter-domain loss, and the text-to-text autoencoder loss with tunable hyperparameter speech-to-text ratio $\beta\in [0,1]$ \cite{CY_cycleGAN-inter-domain-losses},
\begin{align}
\label{eq:my_unpairloss_n}
\begin{split}
\mathcal{L}_{unpair}(f,g,\hat e,d,U) &=\mathcal{L}_{idt}(f,g,\hat e,U)\\&+\beta *\mathcal{L}_{cyc,dom}(f,g,\hat e,d,U)\\&+(1-\beta)*\mathcal{L}_{text}(g,\hat e,d,U)
\end{split}
\end{align}
The identity loss enhances the shared encoder $\hat e(.)$ to preserves important features after translation. The computation of loss in \autoref{fig:illustration-idt-loss} is as follows,
\begin{equation}
    L_{idt}=\rVert \hat{e}(b)-b \rVert_{1} \label{eq:idtloss}
\end{equation}
where the representation is coming from speech $b=f(x)$ or text $b=g(y)$.

The cycle-consistent inter-domain loss is the dissimilarity between the representations of encoded speech and its hypothesis, which aims to let networks learn common knowledge from speech and text. The illustration of loss is shown in \autoref{fig:illustration-cyc-dom-loss} and the definition is as follows,
\begin{equation}\label{eq:cyc_dom}
\begin{split}
     L_{cyc,dom} &= \mathcal{D}(input\_B, cycle\_B)\\
     &= \mathcal{D}(e(x), \hat e(g(d(e(x)))))
\end{split}
\end{equation}
where $\mathcal{D}(.)$ is a distance measure of the distributions. In our previous work, we use Maximum Mean Discrepancy (MMD) because it achieves the best result \cite{CY_cycleGAN-inter-domain-losses}.

The text-to-text autoencoder loss measures a negative log-likelihood that the encoder-decoder network can reconstruct text from unpaired text \cite{text-to-text-autoencoder,karital_inter-domain-loss}, see the orange line in \autoref{fig:architecture-semi-supervised-e2e-asr}. The loss is defined as follows,
\begin{equation}
    L_{text}=-\sum \log \Pr (y|\hat{e}(g(y)))
    \label{eq:loss_text_autoencoder}
\end{equation}

\subsection{CID solely with external text}
\label{sec:better_cyclegan-inter-domain-losses-with-text-data}
In low-resource settings, acquiring paired data or speech data can be costly. Therefore, this section focus on enhancing the model inexpensively.
In our previous work \cite{CY_cycleGAN-inter-domain-losses}, we train model by CID with an equal amount of unlabeled speech and text. However, training a model by CID without additional unlabeled speech and with only external text (i.e., $U=\{X, Y'\}$) might still gain performance improvements. To validate this hypothesis, Table \ref{tab:result1} presents the evaluation of models on Voxforge German test set. These models are trained by jointly CTC-attention objective on paired data $S = \{X, Y \}$ and by CID on speech from paired data and text from Leipzig German corpus \cite{leipzig_corpora} $U=\{X, Y'\}$ simultaneously. The results demonstrate that CID models trained with 10K/100K/300K lines of external text improve WERs from 63.6\% to 38.6/31.2/30.8\% without involving a language model. Moreover, when evaluated with a language model, the CID model improves WERs from 63.1\% to 36.3/29.4/29.1\%. 
These findings highlight the effectiveness of incorporating CID with external text to enhance the performance of E2E model. It also indicates that the CID allows text to benefit not only the language model (LM) but also the encoder-decoder model.

\begin{table*}[htbp]
    \centering
    \begin{tabular}{lclc}\toprule
      \textbf{Model}&\textbf{supervised ratio} $\alpha$  &\textbf{adapted \autoref{eq:my_unpairloss_n} $\mathcal{L}_{unpair}$} &\textbf{CER(\%)} \\\midrule\midrule
       Baseline\cite{CY_cycleGAN-inter-domain-losses} &&&46.9\\
       MIN-UNPAIR-LOSS&0.5  &$\min_{\beta \in \{0,0.1,0.2,...,1.0\}}\mathcal{L}_{unpair}$& \hl{30.6} \\
       MAX-UNPAIR-LOSS&0.5  &$\max_{\beta \in \{0,0.1,0.2,...,1.0\}}\mathcal{L}_{unpair}$& 39.5 \\ 
       AVG-UNPAIR-LOSS&0.5  &$\overline{\mathcal{L}_{unpair}}$& 50.6 \\
       MED-UNPAIR-LOSS&0.5  &Median($\mathcal{L}_{unpair}$)&  50.4 \\\midrule
       DECAY-MIN-UNPAIR-LOSS&decay &$\min_{\beta \in \{0,0.1,0.2,...,1.0\}}\mathcal{L}_{unpair}$&  \hl{29.6}\\
       DECAY-MAX-UNPAIR-LOSS&decay &$\max_{\beta \in \{0,0.1,0.2,...,1.0\}}\mathcal{L}_{unpair}$&  44.1\\ 
       DECAY-AVG-UNPAIR-LOSS&decay &$\overline{\mathcal{L}_{unpair}}$ &  46.6 \\
       DECAY-MED-UNPAIR-LOSS&decay &Median($\mathcal{L}_{unpair}$)&  30.3\\\bottomrule 
    \end{tabular}
    \caption{This table compares the CERs on the Common Voice Finnish test set of models with or without (1) the supervised ratio decay and (2) automatic speech-to-text ratio tuning. We also observe the same conclusion in six languages test sets from Common Voice and Voxforge.}
    \label{tab:improve_interdomain}
\end{table*}

\subsection{Enhanced CID by incorporating automatic hyperparameter tuning}
\label{sec:better_cyclegan-inter-domain-losses}
\begin{figure*}[htbp]
    \subfloat[Training loss of baseline model]
    {\includegraphics[width=0.45\linewidth]{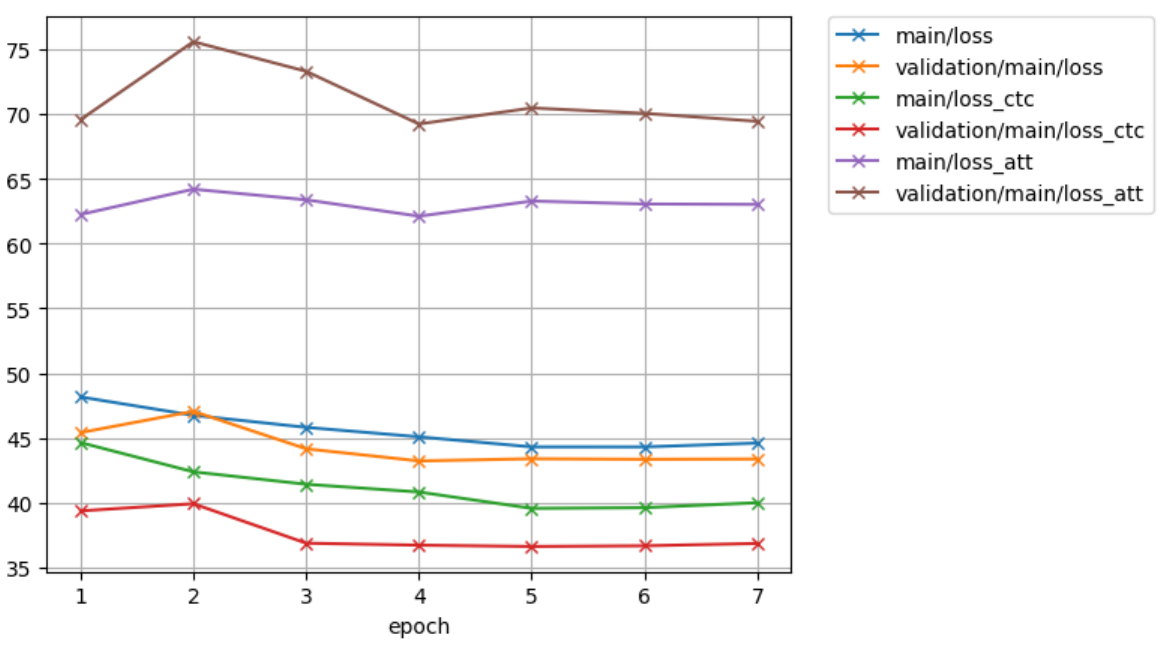}}\hfill
    \subfloat[Accuracy of baseline model]
    {\includegraphics[width=0.45\textwidth]{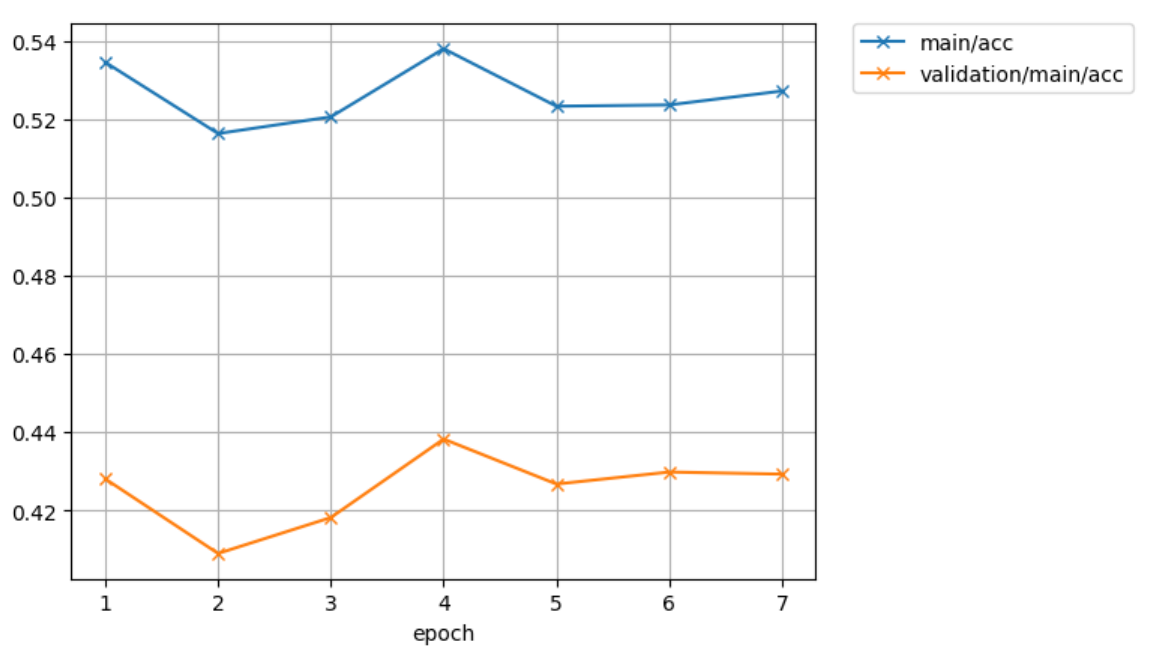}}
    
    \subfloat[Training loss of MIN-UNPAIR-LOSS model]
    {\includegraphics[width=0.45\linewidth]{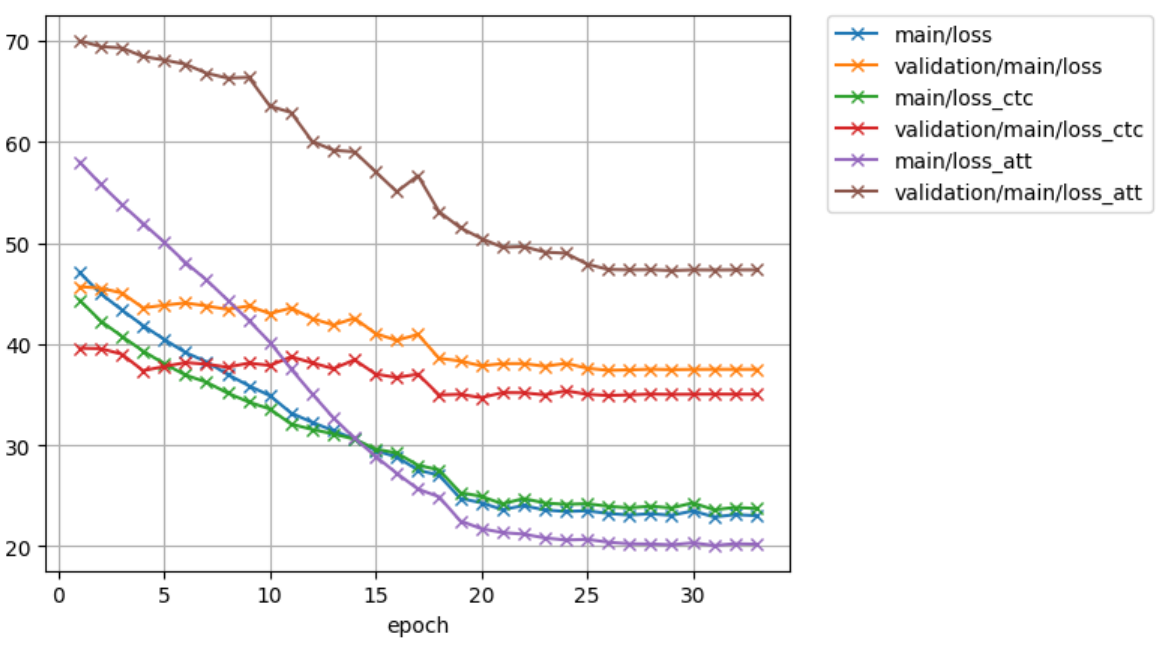}}\hfill
    \subfloat[Accuracy of MIN-UNPAIR-LOSS model]
    {\includegraphics[width=0.45\textwidth]{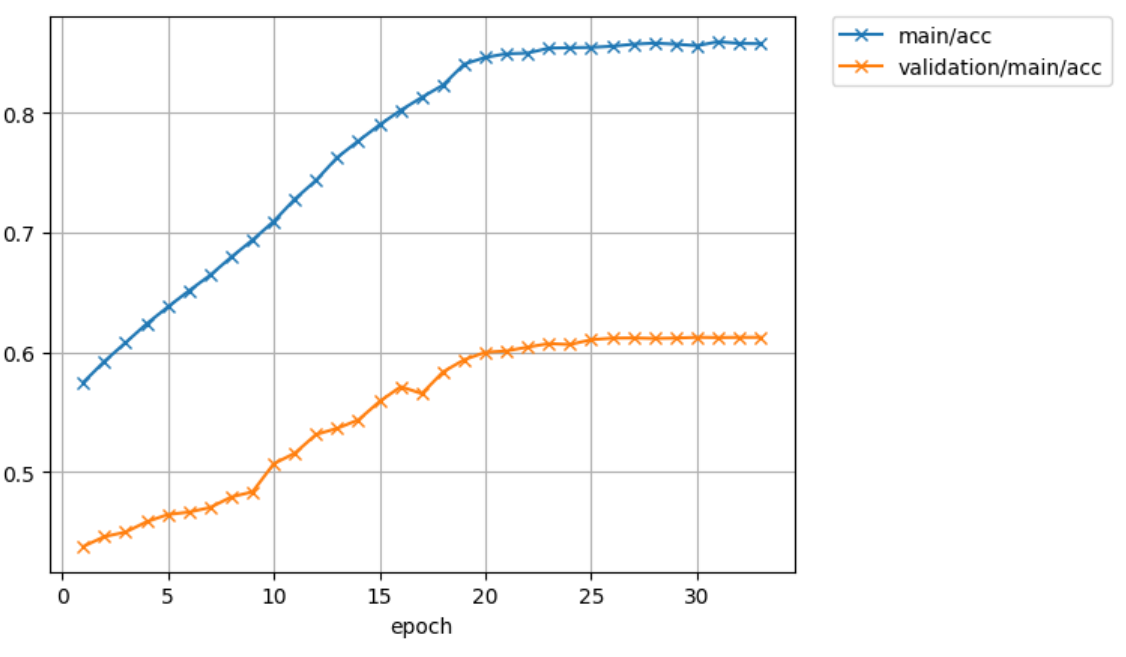}}\hfill
    
    \subfloat[Training loss of MAX-UNPAIR-LOSS model]
    {\includegraphics[width=0.45\textwidth]{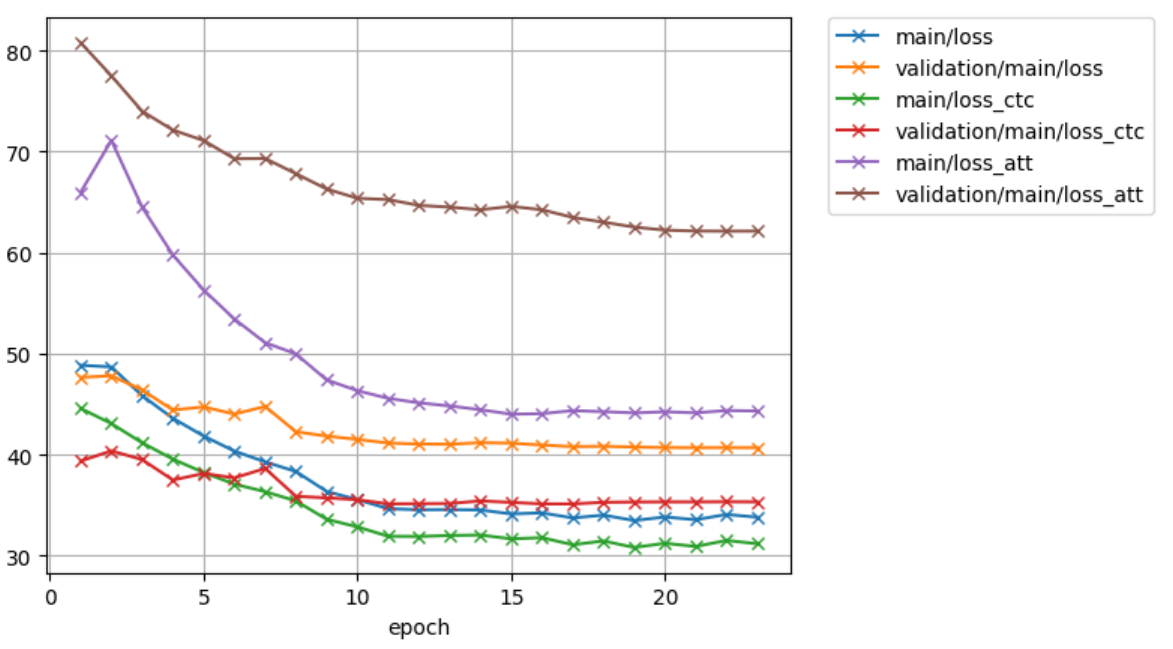}}\hfill
    \subfloat[Accuracy of MAX-UNPAIR-LOSS model]
    {\includegraphics[width=0.45\textwidth] {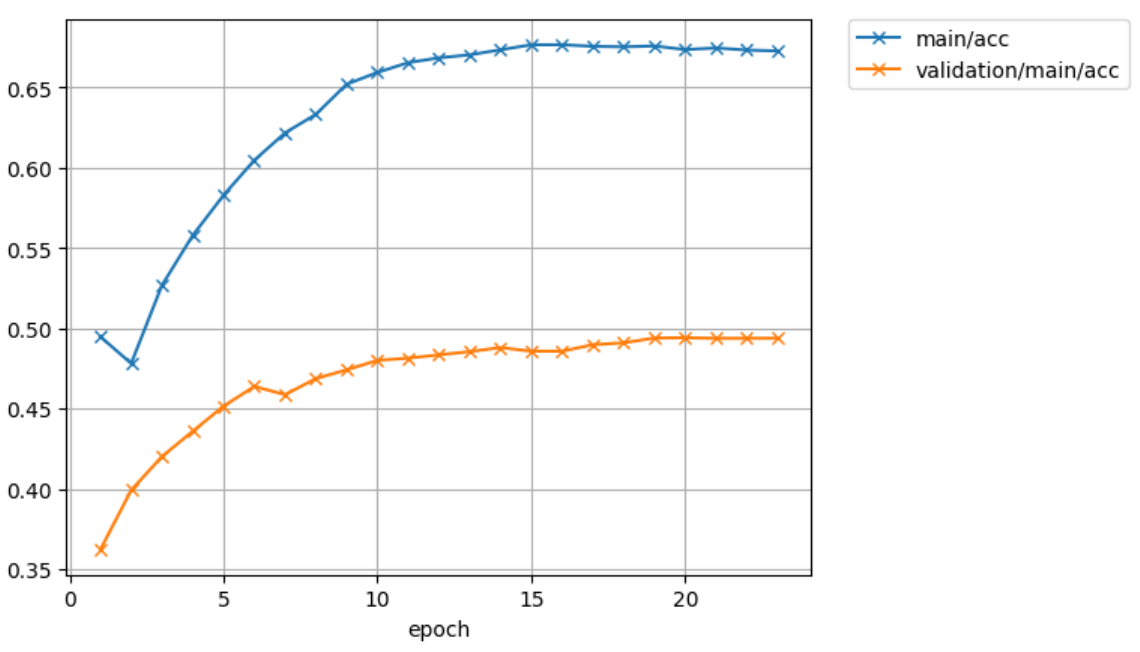}}\hfill
    
    \subfloat[Training loss of AVG-UNPAIR-LOSS model]
    {\includegraphics[width=0.45\linewidth]{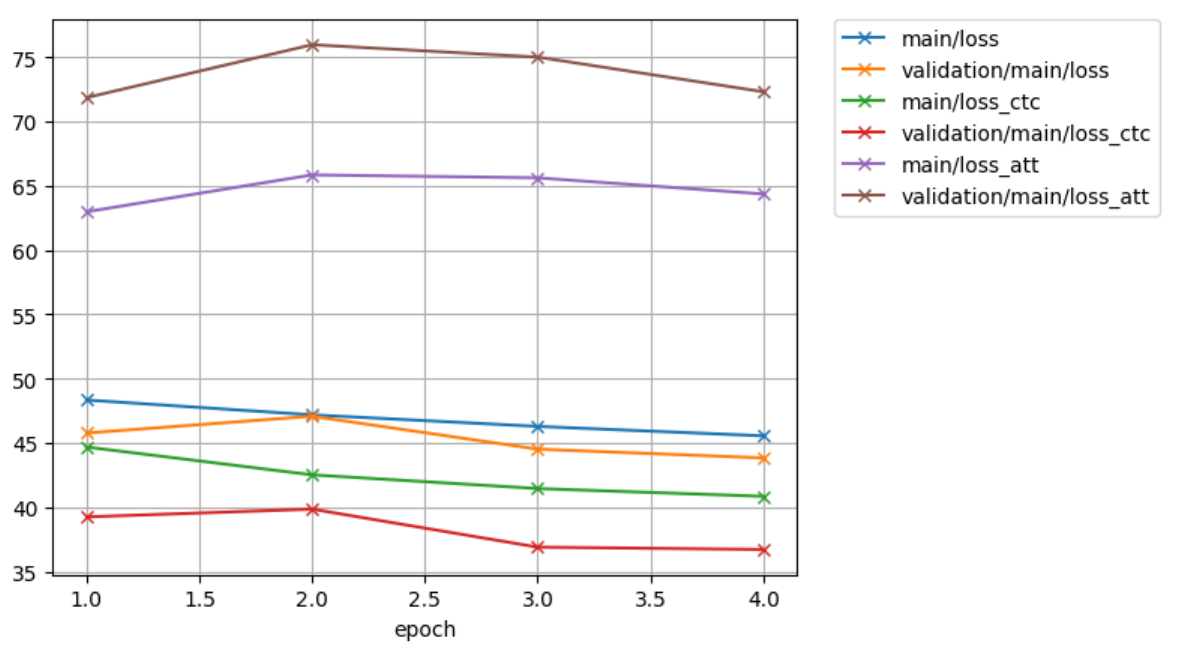}}\hfill
    \subfloat[Accuracy of AVG-UNPAIR-LOSS model]
    {\includegraphics[width=0.45\textwidth]{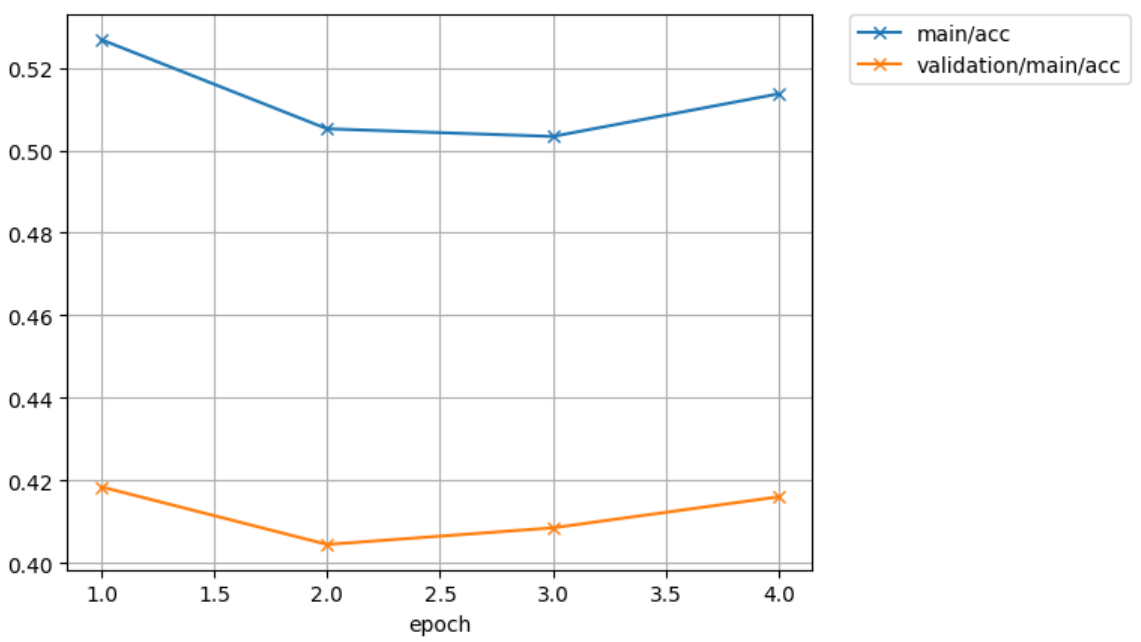}}\hfill
    
    \subfloat[Training loss of MED-UNPAIR-LOSS model]
    {\includegraphics[width=0.45\textwidth]{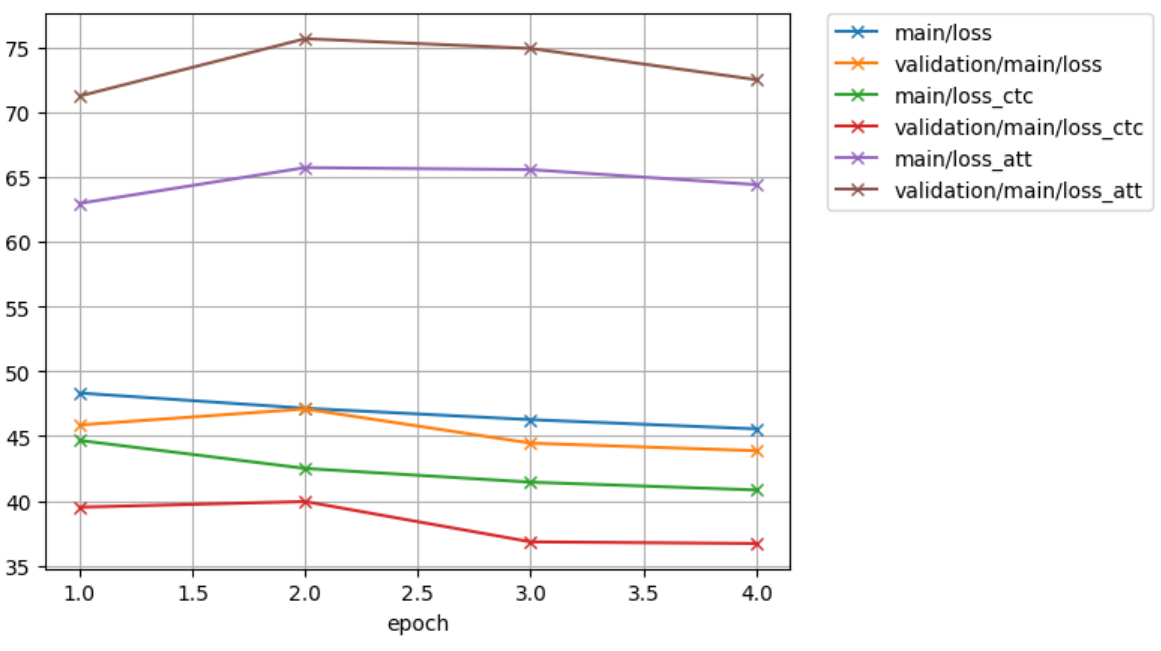}}\hfill
    \subfloat[Accuracy of MED-UNPAIR-LOSS model]
    {\includegraphics[width=0.45\textwidth] {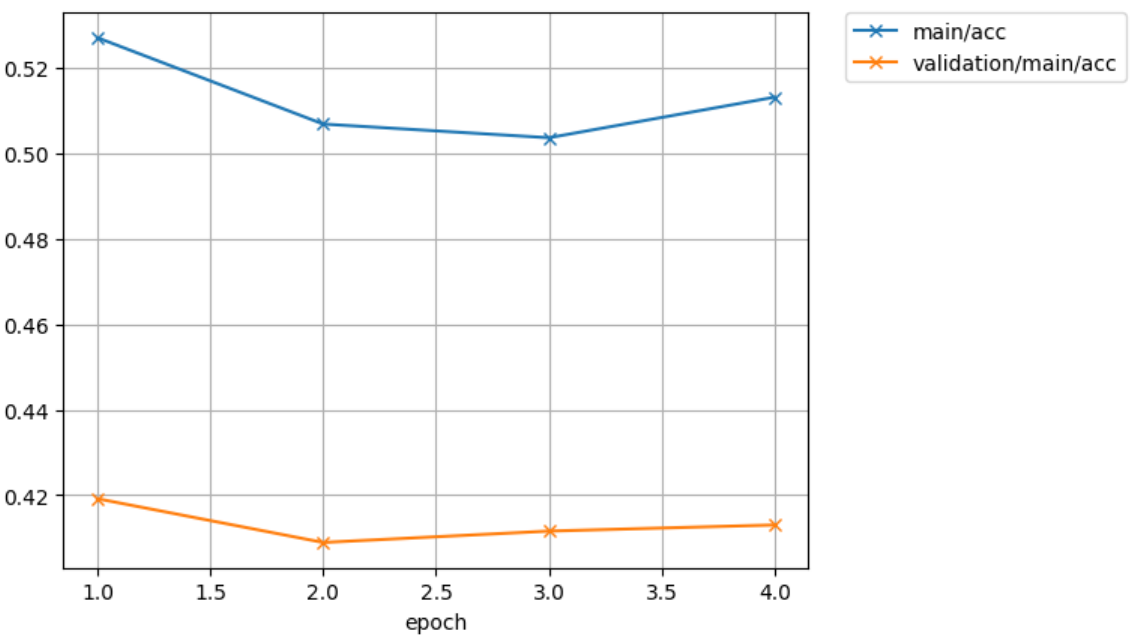}}
    \caption{The training loss (left) and the accuracy (right) of models using different automatic speech-to-text ratio tuning defined in \autoref{tab:improve_interdomain}.}
    \label{fig:training_loss_and_accuracy_different_automatical_hyperparameter_tuning}%
\end{figure*}

\begin{figure*}[htbp]
    \subfloat[Training loss of DECAY-MIN-UNPAIR-LOSS model]
    {\includegraphics[width=0.45\linewidth]{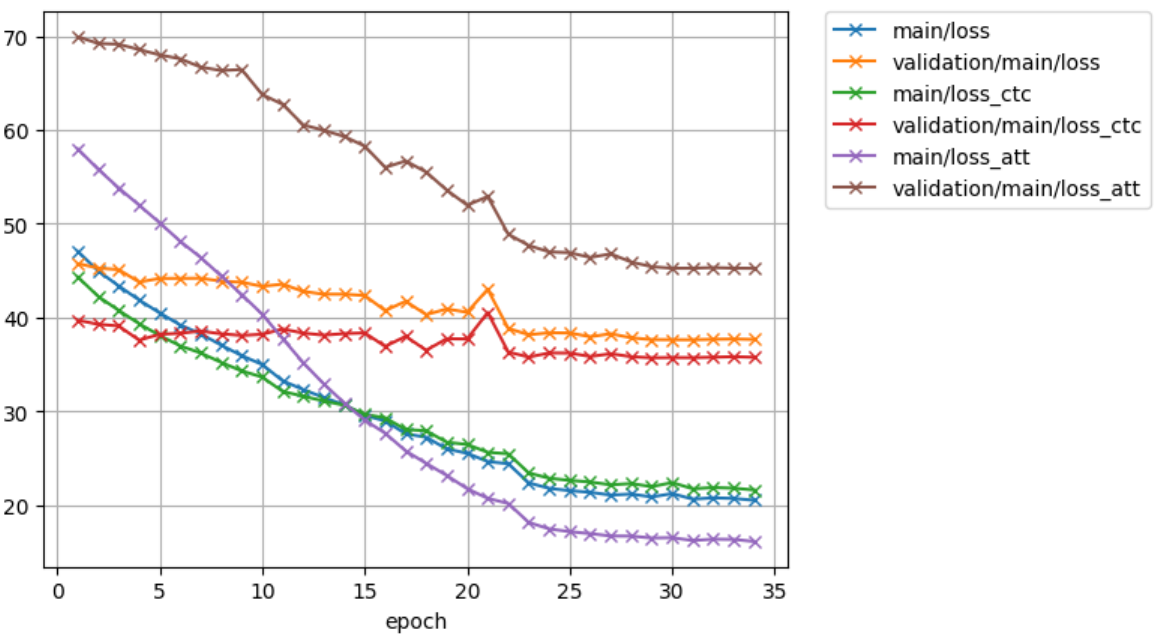}}\hfill
    \subfloat[Accuracy of DECAY-MIN-UNPAIR-LOSS model]
    {\includegraphics[width=0.45\linewidth]{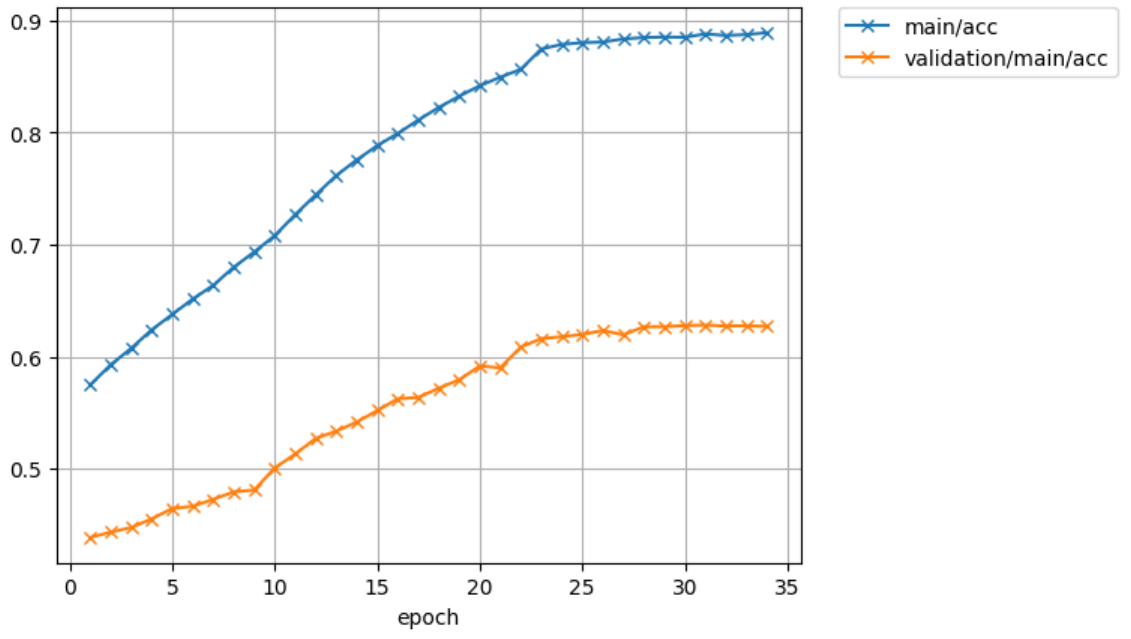}}\hfill
    \subfloat[Training loss of DECAY-MAX-UNPAIR-LOSS model]
    {\includegraphics[width=0.45\textwidth]{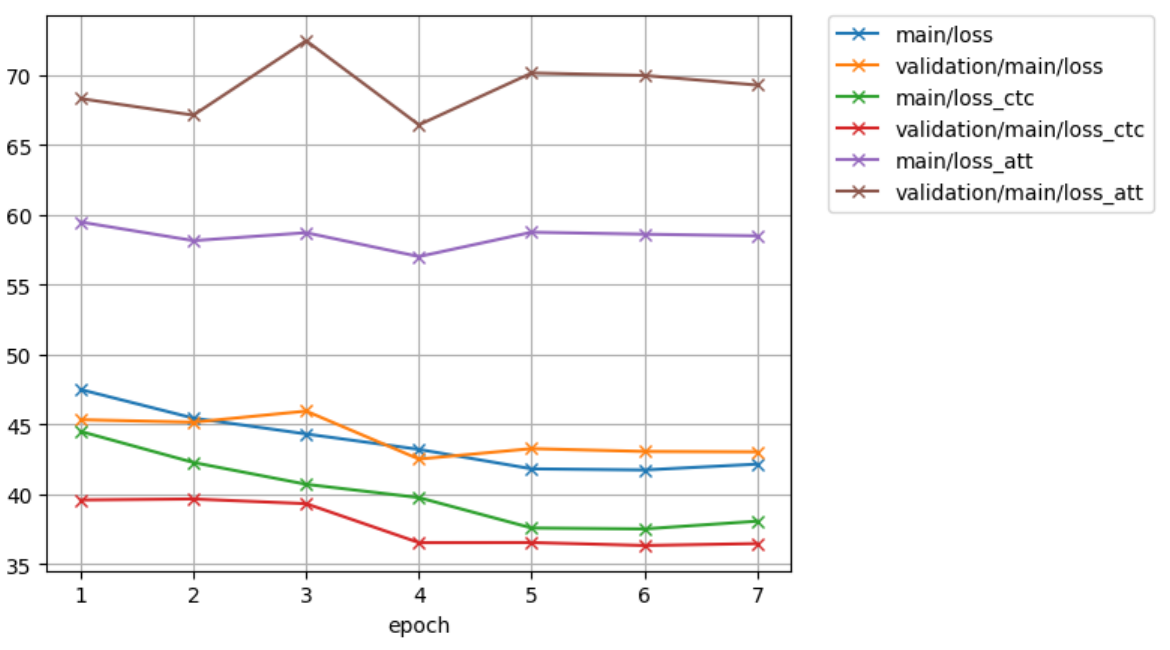}}\hfill
    \subfloat[Accuracy of DECAY-MAX-UNPAIR-LOSS model]
    {\includegraphics[width=0.45\textwidth]{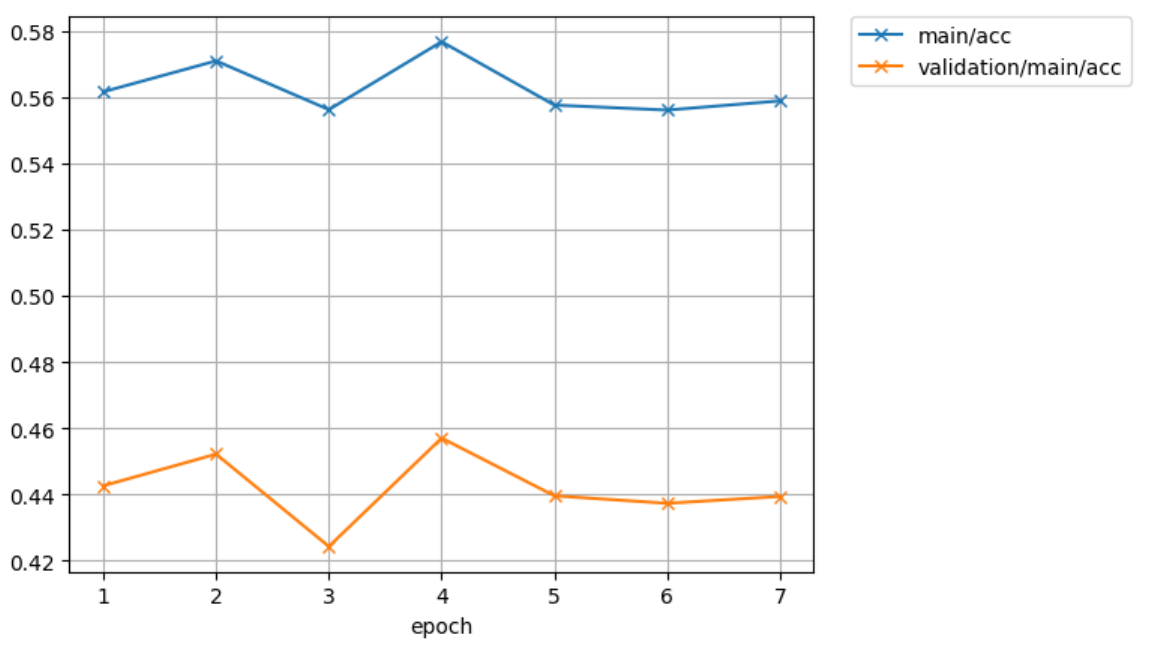}}\hfill
    \subfloat[Training loss of DECAY-AVG-UNPAIR-LOSS model]
    {\includegraphics[width=0.45\textwidth]{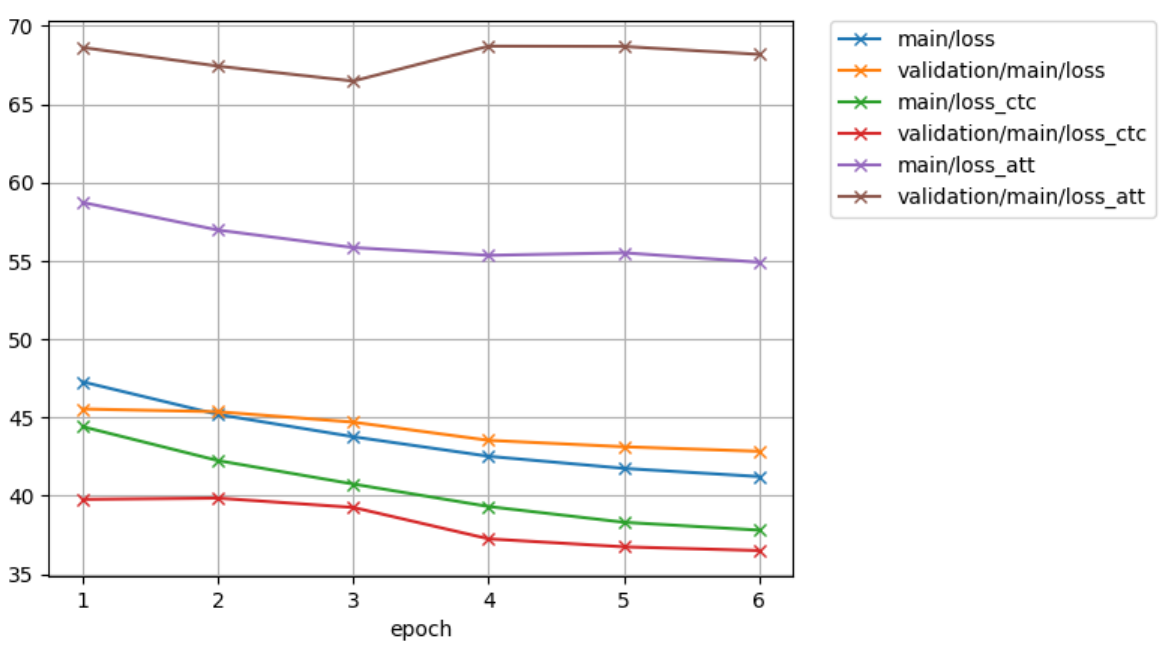}}\hfill
    \subfloat[Accuracy of DECAY-AVG-UNPAIR-LOSS model]
    {\includegraphics[width=0.45\textwidth]{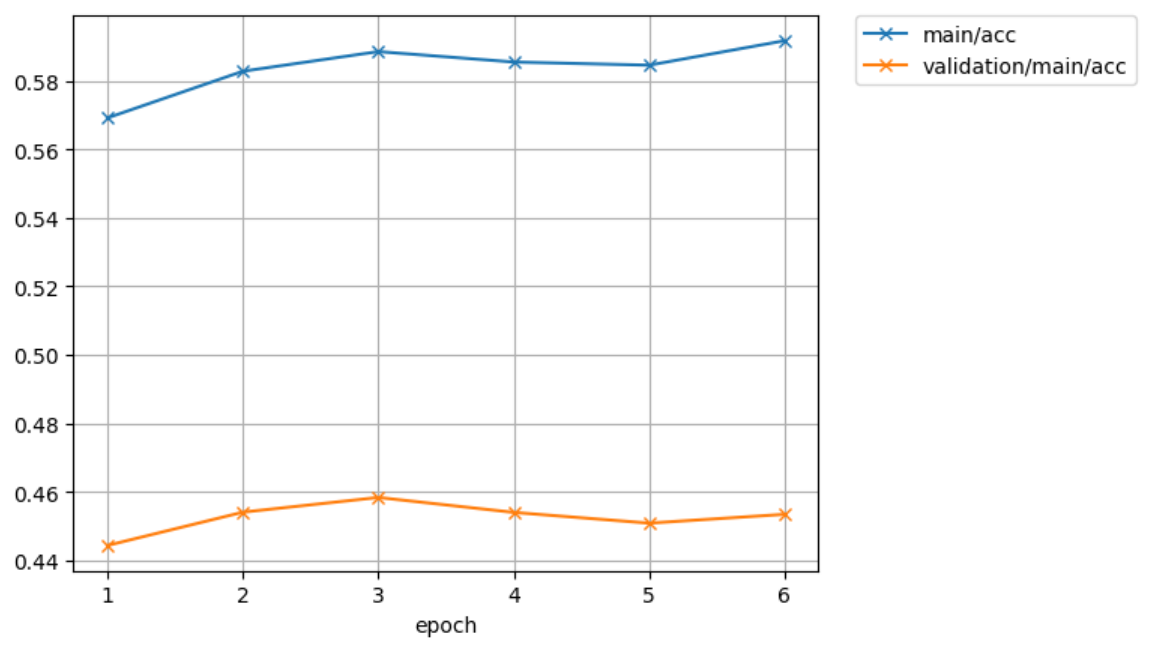}}\hfill
    \subfloat[Training loss of DECAY-MED-UNPAIR-LOSS model]
    {\includegraphics[width=0.45\textwidth] {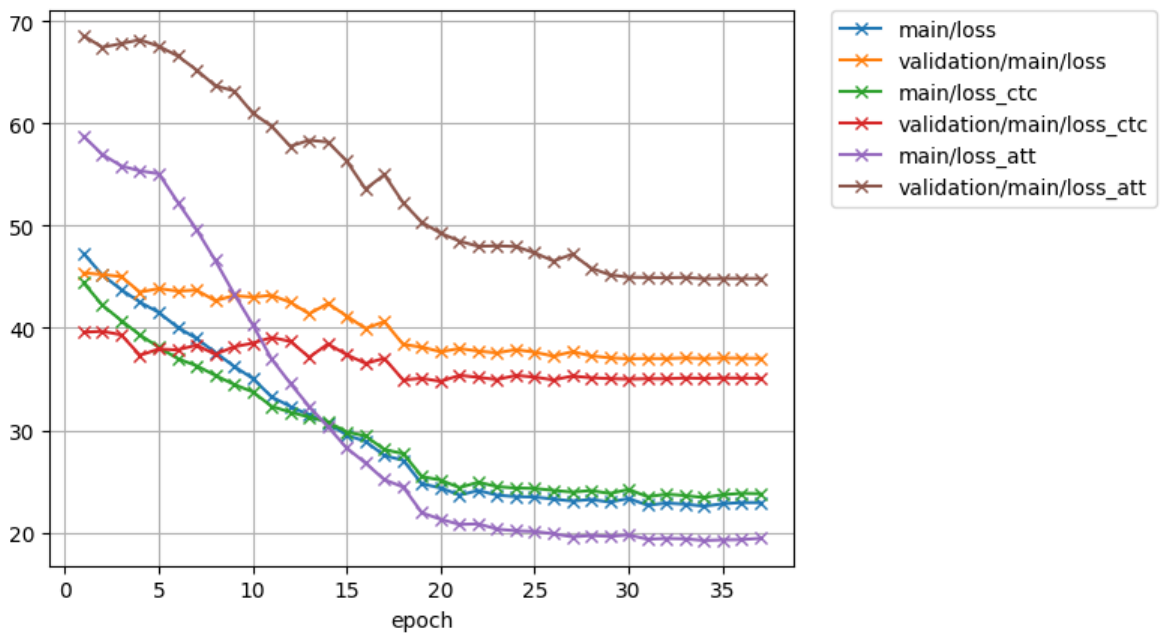}}\hfill
    \subfloat[Accuracy of DECAY-MED-UNPAIR-LOSS model]
    {\includegraphics[width=0.45\textwidth] {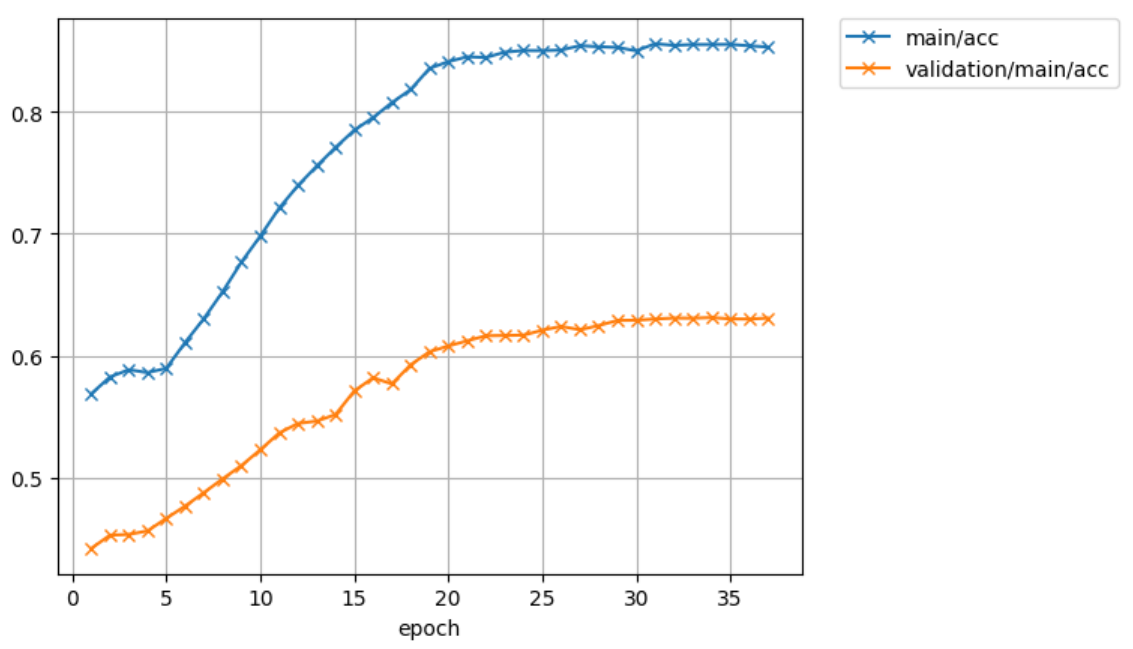}}
    
    \caption{The training loss and accuracy of models using supervised ratio decay and different automatic speech-to-text ratio tuning defined in \autoref{tab:improve_interdomain}.}
    \label{fig:training_loss_and_accuracy_different_automatical_hyperparameter_tuning_and_supervised_ratio_decay}%
\end{figure*}

Although the CID model achieves a significant reduction in character error rate (CERR) across English datasets, WSJ and Librispeech, as well as low supervision non-English datasets (Voxforge) \cite{CY_cycleGAN-inter-domain-losses}, it requires effort to tune the two hyperparameters, the supervised ratio $\alpha$ and the speech-to-text ratio $\beta$, for each dataset. To streamline the training pipeline, we propose using supervised ratio decay and automatic speech-to-text ratio tuning by performing an operation on the unsupervised losses with all the possible values for the speech-to-text ratio during the training. The details are as follows: Firstly, we suggest that the model obtains lots of guidance from the supervision data at the early stages of training. Therefore, $\alpha$ starts at 0.9 for the first three epochs and gradually decays after three epochs until the training is completed, which enables the model to explore the unpaired data with increased flexibility. Secondly, we integrate the speech-to-text ratio into the training process, we propose to use minimal, maximal, average, or median operations on the unsupervised losses with $\beta$ from $0.0$ to $1.0$. Table \ref{tab:improve_interdomain} shows our proposed adapted unsupervised losses and the corresponding CERs on the Common Voice Finnish test set. This table reveals that the model using minimal operation outperforms the ones using other operations and baseline. The best model is the model using the supervised ratio decays and minimal operations on the unsupervised losses over $\beta$. We observe the same conclusion in six languages from Common Voice and Voxforge. \autoref{fig:training_loss_and_accuracy_different_automatical_hyperparameter_tuning} and \autoref{fig:training_loss_and_accuracy_different_automatical_hyperparameter_tuning_and_supervised_ratio_decay} present the training loss and the accuracy of baseline and models trained by our adapted objective in \autoref{tab:improve_interdomain}. The model using minimal operation on unsupervised loss performs stable and improved accuracy during the training, whereas the baseline and other models using maximum, average, and median operations produce mismatched training loss and validated loss, as well as fluctuating model accuracy during the training. These figures resonated with the result from the \autoref{tab:improve_interdomain}, the model trained by \autoref{eq:semie2e} using supervised ratio decay and performing minimal operation on unsupervised loss achieves the best performance.

\begin{figure*}[htb!]  
\begin{tikzpicture}
\begin{axis}[%
width=0.5\textwidth,
height=0.3\textwidth,
xlabel={Model generation (Finnish)},
xtick={0,1,2,3,4,5,6,7,8},
xticklabels={$M_0$,$M_1$,$M_2$,$M_3$,$M_4$,$M_5$,$M_6$,$M_7$,$M_8$},
ytick={10,20,30,40,50,60,70,80,90},
ylabel={WER(\%)}
]
   \addplot coordinates{
        (0, 77.4)
        (1, 60.1)
        (2, 58.9)
        (3, 57.0)
        (4, 59.2)
        (5, 58.4)
        (6, 55.5)
        (7, 57.5)
        (8, 55.1)
   };\addlegendentry{Baseline (NST)};
    \addplot coordinates{
        (0, 77.4)
        (1, 56.8)
        (2, 51.4)
        (3, 55.1)
        (4, 50.7)
        (5, 48.4)
   };\addlegendentry{cNST};
   \end{axis}
\end{tikzpicture}%
\begin{tikzpicture}
\begin{axis}[%
width=.5\textwidth,
height=.3\textwidth,
xlabel={Model generation (Greek)},
xtick={0,1,2,3,4,5,6,7},
xticklabels={$M_0$,$M_1$,$M_2$,$M_3$,$M_4$,$M_5$,$M_6$,$M_7$},
ytick={10,20,30,40,50,60,70,80,90},
ylabel={WER(\%)}
]
   \addplot coordinates{
        (0, 63.2)
        (1, 49.2)
        (2, 41.1)
        (3, 37.1)
        (4, 36.1)
        (5, 35.8)
        (6, 34.8)
        (7, 34.0)
   };\addlegendentry{Baseline (NST)};
    \addplot coordinates{
        (0, 65.6)
        (1, 42.2)
        (2, 34.0)
        (3, 31.3)
        (4, 34.4)
        (5, 31.4)
        (6, 29.4)
        
   };\addlegendentry{cNST};
   \end{axis}
\end{tikzpicture}
\captionof{figure}{WERs on the Common Voice (Finnish and Greek) test set against model generations.}\label{fig:wer_model_gens2}
\end{figure*}
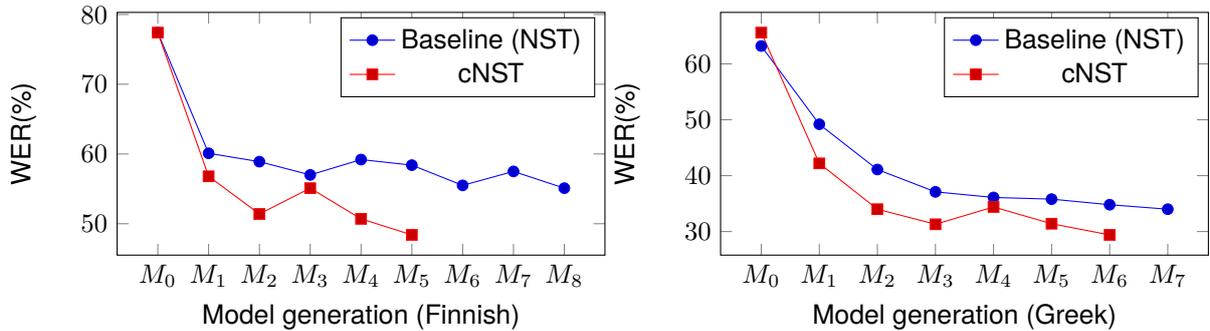

\subsection{Noisy student training with CycleGAN and inter-domain losses (cNST) for low-resource languages} 
\label{sec:cyclegan-inter-domain-losses_and_NST}
NST for speech recognition is effective when sufficient paired data is available. However, the paired data and unlabeled speech are often limited in a low-resource setting. That leads to a low performance teacher model, which generates low-quality labels for unlabeled speech; the training for the student model can be severely affected, resulting in inefficient training.

We aim to improve the teacher model with little effort and less cost regarding time and finances. \autoref{sec:better_cyclegan-inter-domain-losses-with-text-data} demonstrates that the model can be improved by CID solely with external text. Therefore, we propose to exploit the enhanced CID in \autoref{sec:better_cyclegan-inter-domain-losses} and external text to improve the teacher model. A LM is also trained with the in-domain and external text $\{Y, Y'\}$. The NST algorithm is revised as follows,
\begin{enumerate}
    \item Train $M_0$ on $S$ using SpecAugment.
    \item \ul{Train $M_1$ on $S$ and $U=\{Y'\}$ by enhanced CID and using SpecAugment. Set $M=M_1$}.
    \item Fuse $M$ with LM and measure performance. 
    \item Generate labelled dataset $M(X')$ with fused model.
    \item Mix dataset $M(X')$ and $S$. Use mixed dataset to train new model $M'$ with SpecAugment.
    \item Set $M=M'$ and go to 3.
\end{enumerate}
where the initial model $M_0$ is trained with the paired data $S$ using SpecAugment \cite{specAug1}, and we further re-train it at the stage 2 using the enhanced CID with external text with SpecAugment. At stage 3, the teacher model is then fused with a LM to generate labels for the unlabeled speech. Subsequently, the student model is iteratively trained with the paired and newly labeled speech data by the supervised objective. We work with small data, so it is better to utilize the available data wisely rather than removing any of it. Therefore, we simplify the NST training recipe, making it easily applicable to all languages by discarding the sophisticated filtering and balancing stages in \cite{Park_NST}.

\begin{table*}[htbp]
    \centering
    \footnotesize
    \begin{tabular}{lcccccc}
        \toprule
          \textbf{Model}   &  \multicolumn{3}{c}{\textbf{Voxforge (WER\%)}}&\multicolumn{3}{c}{\textbf{Common Voice (WER\%)}}\\
               &German &Italien &Dutch& Hungarian & Finnish & Greek\\\midrule\midrule
         Initial Model ($M_0$)&63.1   &71.2    &63.1 &84.8     &77.4 &63.2\\
         Baseline (NST)        &49.7    &47.1    &  58.2  &72.0    &55.1 &34.0\\
         Proposed Method (cNST)& \textbf{27.3}&\textbf{42.0}&\textbf{56.3}   &\textbf{58.6}     &\textbf{48.4} &\textbf{29.4}\\\midrule
         WERR \% (NST-cNST)/NST &45.1 &10.8&3.26&18.6&12.7&13.5\\\bottomrule
    \end{tabular}
        \caption{WERs comparison between baseline best student model and our proposed cNST best student model across corpus. }
    \label{tab:cNST_effectiveness_across_corpus}
\end{table*}
\section{Experimental Setup}
\label{sec:experimental_setup}
\subsection{Dataset}
Common Voice is a massively multilingual collection of transcribed speech, which is also recorded by user on Mozilla website, and recently it reaches 100 languages \cite{common_voice}. We conducted experiments on a subset of European languages which has limited data: Hungarian, Finnish and Greek. Additionally, we ensured that there were no overlapping sentences or speakers between the train, development and test set. The data size of train/development/test sets are in an 80:10:10 ratio and the test set contains at least two hours speech. The train set is further split to five hours paired data and the remaining portion (around three hours to five hours) is dedicated to the unlabeled speech. 
Voxforge consists of user submitted audio clips using their own microphone \cite{Voxforge.org} and has eight European languages. Each language has limited size of data, ranging from approximately eight to twenty hours. In this paper, we evaluate our proposed method on German, Italian and Dutch languages. The train set is further divide into five hours paired data, while the remaining portion is dedicated to the unlabeled speech $X'$.
The Leipzig corpus, which consists of annual collections of documents from various sources such as wikis, news, and the web \cite{leipzig_corpora}, is used as external text $Y'$ in the experiment.

\subsection{Network architecture}
The semi-supervised E2E model using CycleGAN-inter-domain losses is implemented under Espnet1 \cite{espnet} and \cite{CY_cycleGAN-inter-domain-losses}. The model consists of three layers of Vgg \cite{vgg} bidirectional long short-term memory with projection (Vggblstmp) encoder and attention based decoder, which is one layer long short-term memory (LSTM) with 320 units. The text embedding $g(.)$ encodes the labels over $\{Y, Y'\}$ to an one-hot vector and process it by one layer bidirectional long short-term memory (BLSTM). Byte pair encoding (BPE) \cite{bpe_algo,bpe_nmt} is used for some languages, some have better performance without using BPE. The input acoustic feature is 80-bin log-Mel filterbank with three pitch coefficients. For decoding, we use a beam search algorithm with beam size of 20. 
Our training recipe and code\footnote{\url{https://github.com/chiayuli/Improved-NST-for-low-resource-language.git}}
\begin{table*}[htbp]
\small
    \centering
    \begin{tabular}{lll}
    \toprule    
    \textbf{Models}        && \textbf{Hypothesis}\\\midrule
    Ground-Truth     && es ist sehr beständig gegen witterungseinflüsse und insektenbefall \\
    Initial Model && es ist sehr BESTÄNDIGEN ***** WEITEREN SPÄTEREN SECKER\\
    Baseline(NST) && es ist sehr BESTÄNDIGEN ***** WEITEREN EINFLÜSSE *** ************* \\
    CID && es ist sehr BESTÄNDE gegen WEITERUNGSFLÜSSE und IN SEKTEN BEFALL\\
    cNST&& es ist sehr BESTÄNDE gegen WEITERUNGSEINFLÜSSE und INSEKTEN BEFALL\\\midrule
    Ground-Truth && der anspruch ist von der Frau auf den Mann Übergegangen \\
    Initial Model && der SPRUCH ist *** *** **** *** *** VOLLKOMMEN REGELT\\
    Baseline(NST) && der anspruch ist *** *** **** *** *** **** FREI\\
    CID && ER EINE SPRUCH   ist von der frau auf DIE LANDEN Übergegangen\\
    cNST&& der anspruch ist von der frau auf DIE LANDEN Übergegangen\\\midrule
    Ground-Truth && der Traffic des ersten anbieters wird zum zweiten anbieter weitergeleitet \\
    Initial Model && der ******* *** ****** DRITTES   SPÄTER NETZwerK KANN    NETZwerK GELEITET\\
    Baseline(NST) && der TRITTE  IST ALS    anbieters **** *** ZWEI    LIETER   GELEITET\\
    CID && der TRÄFT  IST ES     ANBIETS   werT ZU  zweiten anbieter \hl{WEITER} GELEITET \\
    cNST&& der TRÄFT IST ES anbieters wird ZU zweiten anbieter \hl{WEITER} GELEITET\\
    \bottomrule
    \end{tabular}
     \caption{\label{tab:asr_outpu_example}The hypothesis of all the models on the unlabeled speech from Voxforge German. Note that the words in uppercase are incorrect compared to the ground-truth and the words in yellow means insertion.}
\end{table*}
\begin{table}[htbp]
    \centering 
    \footnotesize
    \begin{tabular}{lcccc}\toprule
        \textbf{Models}         &\textbf{WER(\%)}&\textbf{INS}&\textbf{DEL}&\textbf{SUB}\\\midrule\midrule
        Initial Model           &63.1&1.8&\textcolor{red}{20.6} &40.7\\
        Baseline         &49.7&1.0&\textcolor{red}{21.0}&27.9\\
        CID                     &\textbf{29.4}&\textcolor{red}{3.3}&\textbf{4.0}&22.0\\
        cNST  &\textbf{27.3}&3.2&3.6&\textbf{20.5}\\\bottomrule
    \end{tabular}
    \caption{\label{tab:52_sub_del_ins_wer}The WER, insertion, deletion, and substitution at word level on the Voxforge German test set. Note that all the results are with the same LM.}
\end{table}

\section{Result}
\label{sec:result}
\subsection{WERs against model generation}
\autoref{fig:wer_model_gens2} shows WERs on the Common Voice (Finnish and Greek) test sets against model generations . We trained the models using our proposed algorithm cNST in \autoref{sec:cyclegan-inter-domain-losses_and_NST} and evaluated the teacher model and all the student models at different stages.   
Based on the observed trend in model performance, it is evident that the red line (cNST) demonstrates a steeper progression compared to the blue line (NST) from $M_0$ to $M_1$. This suggests that the enhanced CID plays a crucial role in accelerating the iterative training process and achieving better results compared to the baseline for all the model generations. Besides, red and blue lines fluctuate over the generations, which might be because the models are over-fitting on the train set, but it does not hurt the subsequent student model performance.

\subsection{cNST effectiveness across corpus}
\autoref{tab:cNST_effectiveness_across_corpus} presents the performance of our proposed method, cNST, across various corpora. We examine the baseline best student model and our proposed cNST best student model on Voxforge German, Italien, Dutch and Common Voice Hungarian, Finnish Greek datasets. The result shows that cNST outperforms the baseline by achieving at least $10$\% WERR for most languages. Moreover, when the initial model performs poorly (above 70\% WER), our proposed cNST successfully reduces the WERs to 40$\sim$50\%, indicating the effectiveness of our proposed method. 
\section{Analysis}
\label{sec:analysis}
\subsection{Recognition output}
We want to gain insights and the reasons for the improvements brought about by enhanced CID. Table \autoref{tab:52_sub_del_ins_wer} presents the WER, insertion, deletion, and substitution on the test set of Voxforge German. The initial model experiences a high number of deletion errors, which are propagated to the subsequent student models in the baseline (NST). However, with enhanced CID, the deletion errors decrease from 20.6 to 4.0. On the other hand, there is a side-effect as the insertion errors increase from 1.8 to 3.3. Overall, the subsequent student model of our proposed cNST achieve the best WER and better substitution and deletion.
\subsection{Cherry-pick hypothesis}
Some cherry-pick examples in \autoref{tab:asr_outpu_example} demonstrate that the initial model and baseline experience high deletion errors. However, the baseline exhibits a further worsening of these errors as the student model undergoes iterative training using labels that contain such errors. This observation resonates with the findings presented in \autoref{tab:52_sub_del_ins_wer}. The enhanced CID model and our proposed cNST successfully reduce deletion errors. However, there is still room for improvement in terms of substitution and insertion errors. Interestingly, In the last example, if we combine both insertion words \enquote{WEITER GELEITET} to \enquote{WEITERGELEITET}, it aligns with the correct word in the reference. The issue with insertions can be attributed to inaccurate word boundary predictions from our proposed models.

\section{Conclusion}
We enhance the CID by incorporating automatic hyperparameter tuning and propose an improved noisy student training that leverages the enhanced CID for low-resource languages. The enhanced CID accelerates the iterative self-training process by sorely utilizing external text. The results demonstrate the effectiveness of our proposed method cNST across six non-English languages from two datasets, surpassing the baseline by 10\% WER. 

\newpage
\bibliographystyle{lrec-coling2024-natbib}
\bibliography{lrec-coling2024-example}

\end{document}